\documentclass[conference]{IEEEtran}
\IEEEoverridecommandlockouts
\usepackage{cite}
\usepackage{amsmath,amssymb,amsfonts}
\usepackage{algorithmic}
\usepackage{graphicx}
\usepackage{textcomp}
\usepackage{xcolor}
\usepackage{mdframed}
\def\BibTeX{{\rm B\kern-.05em{\sc i\kern-.025em b}\kern-.08em
    T\kern-.1667em\lower.7ex\hbox{E}\kern-.125emX}}

\usepackage{tabularx}
\usepackage{booktabs}
\usepackage[T1]{fontenc}
\usepackage{multirow}
\usepackage{colortbl}
\usepackage{xurl}
\usepackage[hidelinks]{hyperref}

\begin{document}

\title{MDWD: A Street-Level Dataset for Municipal Solid Waste Detection in Dense Urban Environments
\thanks{This work was conducted within the \href{https://www.um.edu.mt/research/dawl/}{DAWL AI Lab} at the University of Malta's Department of Artificial Intelligence, as part of the project \textit{Application of AI and Computer Vision to Optimise Cleansing Operations in Malta (AICOM)}, financed by the Government of Malta's Cleansing and Maintenance Division (CMD).}
}

\author{
\IEEEauthorblockN{Andrea Filiberto Lucas}
\IEEEauthorblockA{
University of Malta\\
andrea.f.lucas@um.edu.mt
}
\and
\IEEEauthorblockN{Mark Bugeja}
\IEEEauthorblockA{
University of Malta\\
mark.bugeja@um.edu.mt
}
\and
\IEEEauthorblockN{Carl James Debono}
\IEEEauthorblockA{
University of Malta\\
carl.debono@um.edu.mt
}
\and
\IEEEauthorblockN{Dylan Seychell}
\IEEEauthorblockA{
University of Malta\\
dylan.seychell@um.edu.mt
}
}

\maketitle

\begin{abstract}
Automated visual monitoring of urban environments is a growing Computer Vision research area, but municipal solid waste detection remains under-represented in dedicated benchmark resources. Existing waste-related datasets predominantly address individual litter detection, aerial imagery, or image-level classification, and none simultaneously provide street-level imagery, instance-level localization, and categorization of domestic waste streams within a structured municipal collection context. This paper introduces the Maltese Domestic Waste Dataset (MDWD), a street-level benchmark comprising 3,697 high-resolution images and 11,461 manually annotated instances across five domestic waste categories representative of Malta's municipal collection system. The dataset captures substantial variation in location, illumination, object scale, occlusion, and urban context. To establish reproducible baselines, a cross-architecture benchmark is conducted across multiple generations of the YOLO family and a transformer-based detector. On the test set, RF-DETR-M achieves the strongest overall performance with an mAP\textsubscript{50} of 94.49\% and an F1-score of 93.56\%, whilst smaller-capacity variants maintain competitive accuracy at substantially reduced parameter counts. These results indicate that MDWD supports effective training across both compact real-time detectors and transformer-based models, establishing a benchmark for future research in vision-based municipal waste monitoring.
\end{abstract}


\begin{mdframed}[
    linewidth=0.4pt,
    roundcorner=4pt,
    innertopmargin=6pt,
    innerbottommargin=6pt,
    innerleftmargin=8pt,
    innerrightmargin=8pt
]
\centering
\small
\textit{This paper has been accepted for publication in the\\
14th IEEE European Conference on Visual Information Processing (EUVIP 2026).}
\end{mdframed}

\section{Introduction}
\label{sec:introduction}
Automated visual monitoring of urban environments is an increasingly active area of Computer Vision (CV) research, addressing the operational pressures faced by municipal administrations as cities grow denser and more populous~\cite{urbanproblem,marasinghe2026urbanplanning}. Municipal solid waste (MSW) management exemplifies this pressure: in Malta, municipal waste generation reached 621~kg per capita in 2024 against a European Union average of 517~kg~\cite{eurostat2026municipalwaste}, and traditional monitoring practices based on manual inspection and citizen reporting remain labor-intensive, reactive, and poorly suited to dense urban environments~\cite{adnan2020municipal}.

From a CV perspective, street-level waste monitoring constitutes a challenging instance-level object detection (OD) problem, characterized by frequent occlusion, substantial scale and illumination variation, perspective distortion from heterogeneous capture devices, visually similar waste categories, and long-tailed class distributions~\cite{naufaldihanif2025detectorcomparison}. Modern detection frameworks have demonstrated strong capability under such conditions~\cite{naufaldihanif2025detectorcomparison}, offering the prospect of automatically detecting and categorizing domestic waste from street-level imagery to support monitoring, non-compliance identification, and operational decision-making~\cite{kumar2026gvp}.

Realizing this potential requires annotated datasets reflecting these conditions within an operational collection context. Existing waste-related datasets~\cite{pisani2024soda,bartolo2026aeriallitter,duras2024seaclear,proenca2020taco, MJU-Waste,trashnet,BDW,sukel2023gigo} predominantly address individual litter detection, aerial imagery, marine debris, or image-level recycling classification, and none simultaneously provide street-level imagery, instance-level localization, and categorization into multiple operationally defined domestic waste streams. Malta provides a representative case study through which this gap can be addressed, with a color-coded domestic collection scheme operating within dense, visually cluttered urbanscapes that are representative of the broader challenges facing street-level waste monitoring more generally.

To address this gap, this paper introduces the \textbf{Maltese Domestic Waste Dataset (MDWD)}, a street-level benchmark comprising 3,697 high-resolution images and 11,461 manually annotated instances across five domestic waste categories representative of Malta's municipal collection system. The five categories correspond to distinct streams encountered during routine collection: Mixed Waste denotes residual household refuse in black bags; Organic Waste comprises biodegradable material in small white bags collected under Malta's organic waste scheme; Recyclable Material covers gray or green bags containing paper, cardboard, plastic, and metal packaging; Orange CMD identifies distinctive orange bags associated with Cleaning and Maintenance Division (CMD) operations; and Other Waste captures infrequently occurring items, such as bulky refuse and glass bottles, that account for the long-tail distribution typical of real-world urban environments. To establish reproducible baselines and characterize dataset difficulty, a cross-architecture benchmark is conducted across multiple generations of the YOLO family and the transformer-based RF-DETR framework.

The main contributions of this work are the introduction of MDWD itself, an operationally grounded five-class taxonomy reflecting a real municipal collection scheme, instance-level street-level annotations capturing the visual challenges characteristic of dense urban environments, and a reproducible cross-architecture benchmark establishing baseline performance to support future research in vision-based MSW monitoring.

\section{Relevant Literature}
\label{sec:related_work}
This section presents the literature underpinning the proposed benchmark, covering urban monitoring applications, waste-related CV datasets, and contemporary OD paradigms.

\subsection{Computer Vision for Urban and Environmental Monitoring}
Vision-based monitoring has become a central tool for managing urban environments at scale, supporting applications ranging from infrastructure inspection to traffic and environmental assessment~\cite{urbanproblem,marasinghe2026urbanplanning,ashraf2026urbaninfrastructure}. The appeal of such systems lies in their ability to convert street-level or aerial imagery into spatially structured information that can support routine operational decision-making without continuous manual inspection~\cite{ashraf2026urbaninfrastructure}. More recently, vision-language models (VLMs) have been proposed as a flexible alternative for urban scene interpretation, offering open-vocabulary querying without task-specific retraining~\cite{torneiro2025urbanmonitoring,vlm}. However, current VLM-based approaches remain predominantly evaluative or exploratory, and Torneiro et al.~\cite{torneiro2025urbanmonitoring} explicitly identify the absence of large, annotated, task-specific datasets as a barrier to rigorous benchmarking in this space. This observation generalizes beyond urban monitoring as a whole: regardless of whether the eventual modeling paradigm is a task-specific detector or a general-purpose VLM, progress depends on the availability of annotated data that reflects the operational conditions of the target deployment environment. Within municipal waste monitoring specifically, this dependency is particularly acute, since waste accumulation exhibits considerable visual heterogeneity across geographic regions, collection schemes, and urban morphologies, limiting the transferability of models trained on visually dissimilar source data.

\subsection{Waste-Related Computer Vision Datasets}
The development of waste-related CV datasets has progressed along several largely independent trajectories, each constrained by a different combination of acquisition setting, annotation granularity, and task formulation. Table~\ref{tab:dataset_comparison} summarizes the datasets most relevant to MDWD, while the discussion below synthesizes the limitations that motivate the present contribution.

\begin{table}[t]
\caption{Comparison of Waste-Related CV Datasets and MDWD.}
\label{tab:dataset_comparison}
\centering
\footnotesize
\renewcommand{\arraystretch}{1.15}
\begin{tabular*}{\columnwidth}{@{\extracolsep{\fill}}lccp{4.6cm}@{}}
\toprule
\textbf{Dataset} & \textbf{Images} & \textbf{Classes} & \textbf{Annotation Type \& Context} \\
\midrule
TrashNet~\cite{trashnet} & 2,527 & 6 & Image-level classification of \newline household litter  \\[2pt]
TACO~\cite{proenca2020taco} & 1,500 & 60 & Bounding boxes \& polygons \newline detection of street-level litter \\[2pt]
GIGO~\cite{sukel2023gigo} & 25,000 & 5 & Multi-label image classification of \newline street-level MSW \\[2pt]
SODA~\cite{pisani2024soda} & 829 & 6 & Bounding boxes \& polygons of \newline UAV litter detection \\[2pt]
GVP~\cite{kumar2026gvp} & 5,000 & 2 & Bounding boxes of\newline street-level binary MSW detection \\
\midrule
\textbf{MDWD (Ours)} & \textbf{3,697} & \textbf{5} & \textbf{Bounding boxes of} \textbf{street-level \newline multi-class MSW detection} \\
\bottomrule
\end{tabular*}
\end{table}

Early efforts such as TrashNet~\cite{trashnet} established image-level classification of recyclable materials under controlled, single-object conditions, and have been valuable for demonstrating the feasibility of transfer learning for waste classification~\cite{jain2024mobilenetbags,jain2024resnetbags}. However, the absence of spatial localization and the controlled acquisition setting limit applicability to cluttered, multi-instance street scenes. TACO~\cite{proenca2020taco} advanced the field by introducing object-level annotations for litter detection and segmentation across a diverse set of outdoor scenes; however, its sixty fine-grained litter categories are oriented towards individual discarded items rather than the bagged, stream-categorized waste characteristic of organized municipal collection.

A second trajectory has focused on aerial and UAV-based litter detection. SODA~\cite{pisani2024soda}, notably also collected within the Maltese context, provides bounding-box and polygon annotations for small-object litter detection from UAV imagery captured over garigue terrain. Bartolo et al.~\cite{bartolo2026aeriallitter} provide a comprehensive review of this UAV-based litter detection landscape, synthesizing datasets such as BDW~\cite{BDW}, SODA, and several ground-based collections including MJU-Waste~\cite{MJU-Waste}, and conclude that the field requires more in-the-wild datasets exceeding several thousand annotated images. While this body of work demonstrates substantial progress in small-object and litter detection, the aerial acquisition perspective removes street-level contextual cues, such as bag color, placement pattern, and surrounding urban infrastructure, that are operationally meaningful for municipal collection workflows. BDW~\cite{BDW} and MJU-Waste~\cite{MJU-Waste} further illustrate the diversity of task formulations explored in this space, employing oriented bounding boxes for rotation-invariant bottle detection and RGB-D semantic segmentation for fine-grained object boundaries respectively; while methodologically valuable, neither addresses street-level categorization of domestic waste streams.

A third trajectory addresses street-level municipal monitoring directly. GIGO~\cite{sukel2023gigo} provides vehicle-mounted street-level imagery with multi-label classification of urban garbage categories, offering the closest conceptual alignment with MDWD in terms of acquisition setting; however, its image-level formulation precludes instance-level localization, limiting its use for tasks requiring spatial identification of individual waste accumulations. GVP~\cite{kumar2026gvp} addresses a comparable street-level acquisition setting through fixed-camera monitoring of garbage vulnerable points, and its binary waste/non-waste annotation scheme demonstrates the feasibility of detector-based monitoring for tracking accumulation patterns over time. Its single-class formulation, however, does not support differentiation between distinct waste streams, which is operationally important where municipal collection is organized around separate categories such as mixed, organic, and recyclable waste.

Considered collectively, none of the reviewed datasets simultaneously provide street-level imagery, instance-level bounding-box annotations, and categorization into multiple operationally defined domestic waste streams. MDWD is designed to occupy this position directly, providing street-level imagery annotated at the instance level across five categories that mirror an operational municipal collection scheme.

\subsection{Object Detection Architectures and Benchmarking Considerations}
\label{subsec:LR-OD}
The choice of detection architecture for MSW monitoring reflects a trade-off between accuracy and deployment constraints such as latency and model size. Two-stage detectors such as Faster R-CNN achieve strong accuracy but at a computational cost that limits real-time deployment~\cite{naufaldihanif2025detectorcomparison}, leaving the YOLO family as the dominant real-time paradigm: YOLO11 improves feature extraction through enhanced spatial attention~\cite{yolo11}, YOLO12 adopts an attention-centric design while retaining real-time inference~\cite{tian2025yolov12}, and YOLO26 introduces an edge-oriented design with small-target-aware label assignment and an NMS-free head~\cite{hidayatullah2026yolo26}. Kumar et al.~\cite{kumar2026gvp} report YOLO11m outperforming RT-DETR at a mAP\textsubscript{50} of 0.91 on street-level waste imagery, motivating these YOLO variants as the primary CNN-based reference. RF-DETR offers a complementary transformer-based paradigm, applying neural architecture search to retain global attention beneficial for cluttered, overlapping scenes while achieving real-time inference~\cite{robinson2026rfdetrneuralarchitecturesearch}. Its inclusion alongside the three YOLO generations yields a benchmark that spans CNN-based, attention-centric and transformer-based paradigms under a common protocol.

Standardized, publicly available benchmarks allow comparisons to be attributed to genuine model characteristics rather than dataset-specific artifacts, enabling subsequent research to build on a shared evaluation protocol~\cite{ashraf2026urbaninfrastructure}. Given the absence of an existing dataset satisfying the requirements identified above, establishing such a benchmark constitutes the central motivation for MDWD.

\section{Methodology}
This study adopts a multi-stage methodology comprising dataset construction, annotation, data preparation, model selection, and comparative evaluation. The MDWD is publicly available via Roboflow\footnote{\url{https://universe.roboflow.com/um-dawl-ai-lab/mdwd-maltese-domestic-waste-dataset}}.

\subsection{Dataset Construction and Annotation}
MDWD comprises 3,697 high-resolution street-level images collected across Malta and Gozo between Winter 2024 and late Summer 2025, spanning multiple seasons and meteorological conditions to ensure temporal diversity in illumination and waste presentation patterns. Images were acquired by a third-party team using a variety of consumer-grade mobile devices, predominantly on foot with a smaller proportion captured from moving vehicles. Collection was conducted opportunistically across geographically diverse localities, with each image depicting a distinct waste pile or collection event; burst captures and video-derived frames were explicitly excluded. Images were acquired at a median resolution of $3024 \times 4032$ pixels, with an average of 3.1 annotated instances per image.

\begin{figure}[t]
  \centering
  \includegraphics[width=\columnwidth]{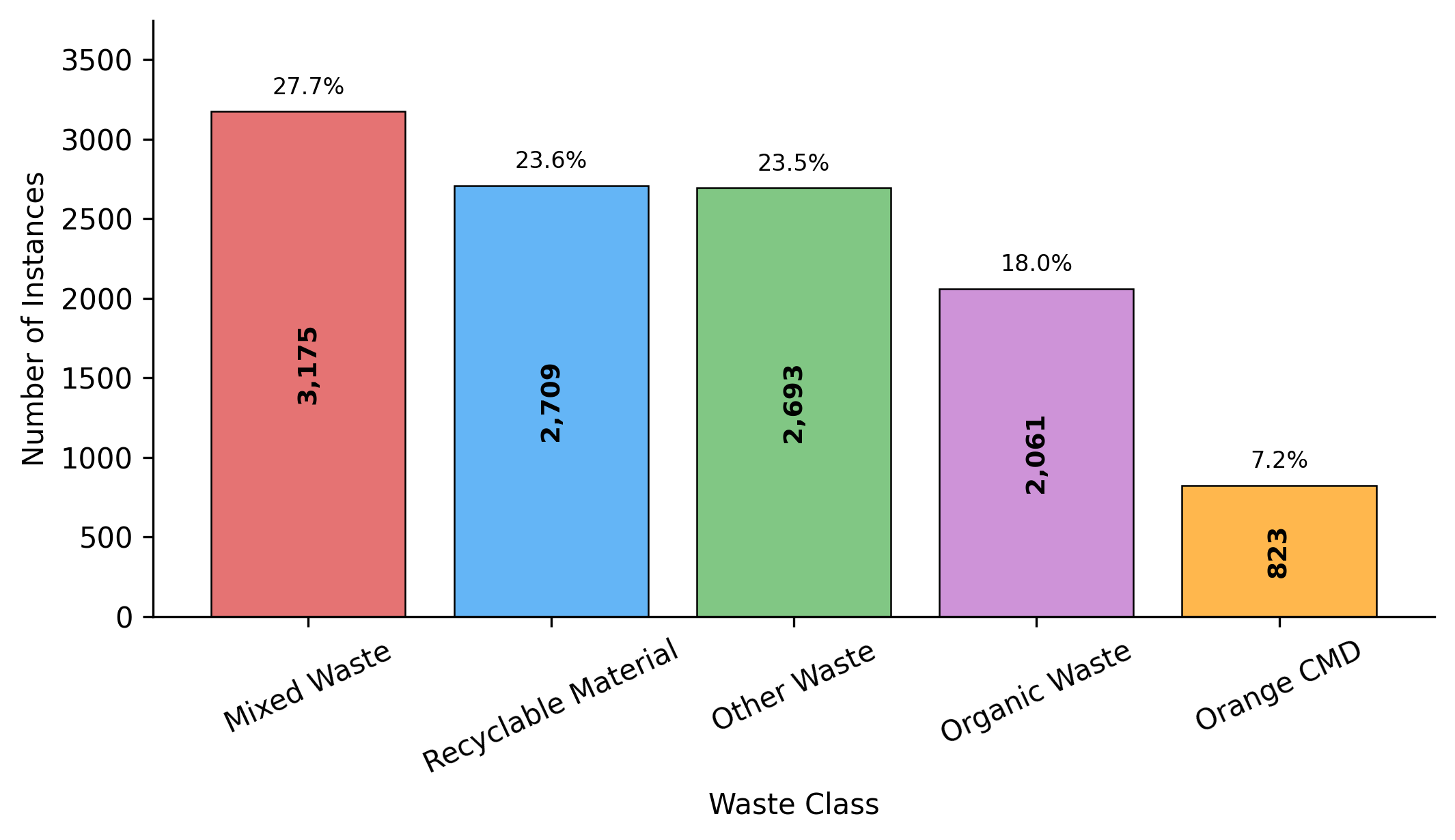}
  \caption{Class distribution of the 11,461 annotated instances in the MDWD. The dataset exhibits moderate class imbalance, with Mixed Waste representing the largest category (27.7\%) and Orange CMD the smallest (7.2\%), reflecting real-world waste occurrence patterns.}
  \label{fig:class_distribution}
\end{figure}

All annotations were produced manually by a team of trained annotators using the Roboflow annotation platform, then reviewed and consolidated by a single experienced lead annotator in a dedicated quality-assurance pass over the full dataset, correcting boundary errors, resolving category disagreements, and removing ambiguous instances. Annotations take the form of axis-aligned bounding boxes fitted tightly to the visible extent of each instance. Objects were annotated if at least approximately 20\% of their extent was visible; no predictive labeling was applied to occluded regions, and instances of indeterminate category were excluded. The five waste categories correspond directly to Malta's municipal collection streams, as summarized in Figure~\ref{fig:class_distribution}, and are described in Section~\ref{sec:introduction}.

The distribution is relatively balanced across the four principal categories. Orange CMD constitutes the least represented class owing to the lower operational frequency of CMD collection activities within the surveyed areas; however, its highly distinctive visual characteristics mitigate inter-class ambiguity.

\subsection{Dataset Preparation and Augmentation}
The dataset was partitioned using an 80/10/10 train--validation--test split, corresponding to 2,958 training, 370 validation, and 369 test images. All images first underwent automatic orientation correction to remove EXIF-encoded rotation artifacts, then were resized to $640 \times 640$ pixels using a stretch-based strategy consistent with the native input dimensions of the evaluated YOLO architectures. Stretch-based resizing was preferred over letterboxing to avoid padding artifacts that interact inconsistently with mosaic augmentation.

\begin{figure}[t]
    \centering
    \includegraphics[width=0.9\columnwidth]{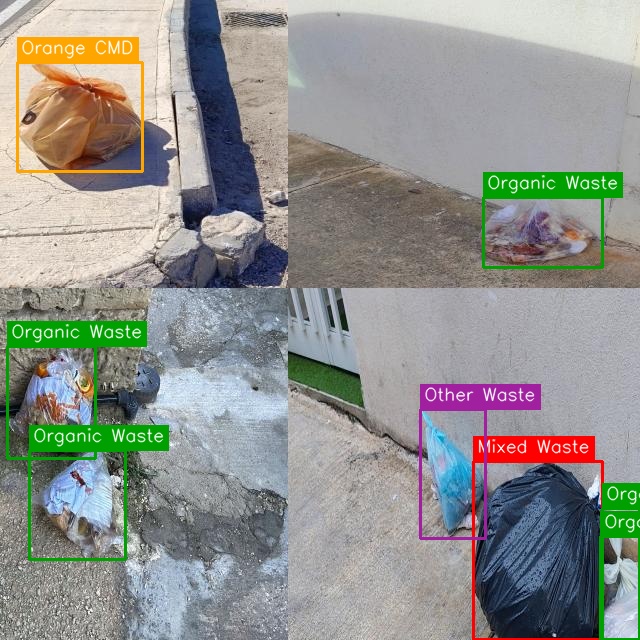}
    \caption{Representative MDWD training sample after offline mosaic augmentation, showing four composited images with color-coded bounding-box annotations across multiple waste categories.}
    \label{fig:dataset_sample}
\end{figure}

Offline augmentation was applied exclusively to the training partition via the Roboflow platform, generating ten augmented variants per image and expanding the effective training set to 29,580 images. Fixing augmentations prior to export improves reproducibility by ensuring a consistent augmented training corpus across all evaluated models. The augmentation pipeline is summarized in Table~\ref{tab:augmentation}. Photometric transformations simulate illumination variability across times of day and sensor types; horizontal and vertical flipping were retained as robustness-oriented geometric transformations; Gaussian blur (up to 1\,px) approximates mild optical defocus; motion blur (up to 40\,px) reflects handheld and vehicle-mounted capture conditions and was selected empirically based on observed validation performance improvement; and $2{\times}2$ mosaic augmentation increases contextual diversity and improves small-object detection by compositing four training images per input.

\begin{table}[t]
  \caption{Offline Augmentation Pipeline Applied Through the Roboflow Platform Prior to Export.}
  \label{tab:augmentation}
  \centering
  \begin{tabular*}{\columnwidth}{@{\extracolsep{\fill}}lr@{}}
    \toprule
    \textbf{Augmentation} & \textbf{Configuration} \\
    \midrule
    Resize                & $640 \times 640$ (Stretch) \\
    Horizontal Flip       & 50\% probability \\
    Vertical Flip         & 50\% probability \\
    Hue Adjustment        & $\pm 19^{\circ}$ \\
    Saturation Adjustment & $\pm 29\%$ \\
    Brightness Adjustment & $\pm 17\%$ \\
    Exposure Adjustment   & $\pm 10\%$ \\
    Gaussian Blur         & Up to 1\,px \\
    Motion Blur           & Up to 40\,px \\
    Mosaic Augmentation   & Enabled ($2{\times}2$) \\
    \bottomrule
  \end{tabular*}
\end{table}

\subsection{Evaluated Models and Training Configuration}
To establish reproducible baselines across representative detection paradigms, three generations of the YOLO family, namely YOLO11, YOLO12, and YOLO26, as well as the transformer-based RF-DETR framework were evaluated, motivated by the demonstrated effectiveness of modern OD architectures for waste and litter detection tasks~\cite{bartolo2026aeriallitter}. YOLO11, YOLO12, and RF-DETR were evaluated at nano (N), small (S), and medium (M) scales; YOLO26 was additionally evaluated at large (L) scale to provide a more extensive capacity analysis of the most recent generation. Detector architecture and model scale constitute the sole independent variables across experiments; all other conditions are held constant to the greatest extent permitted by the respective training environments.

All YOLO experiments were conducted using the Ultralytics framework (v8.4.60), PyTorch 2.10.0, and CUDA 13.0 on an NVIDIA GeForce RTX 4090 with 24\,GB of device memory. Models were initialized from publicly available COCO pre-trained checkpoints, and a fixed random seed of 42 was employed alongside deterministic execution to promote reproducibility. The shared hyperparameter configuration is presented in Table~\ref{tab:hyperparams}, and all runs were tracked via Weights~\&~Biases.

\begin{table}[t]
  \caption{Shared Training Hyperparameters for All Locally Trained YOLO-Family Models.}
  \label{tab:hyperparams}
  \centering
  \begin{tabular*}{\columnwidth}{@{\extracolsep{\fill}}lr@{}}
    \toprule
    \textbf{Parameter} & \textbf{Value} \\
    \midrule
    Image Size                & $640 \times 640$ \textit{(Stretch)} \\
    Epochs                    & 100 \\
    Patience (Early Stopping) & 10 \\
    Batch Size                & 32 \\
    Optimiser                 & AdamW \\
    Initial Learning Rate     & 0.001 \\
    Weight Decay              & 0.0005 \\
    Momentum                  & 0.937 \\
    Warmup Epochs             & 3 \\
    Seed                      & 42 \\
    \bottomrule
  \end{tabular*}
\end{table}

RF-DETR experiments were conducted via the Roboflow cloud training platform, with models initialized from Objects365 pre-trained weights and trained for 100 epochs with the same early stopping criterion; all remaining hyperparameters followed Roboflow platform defaults. While both families were trained on the same offline augmented dataset for the same number of epochs, differences in pre-training source and training infrastructure mean that cross-architecture comparisons should be interpreted as a practically motivated benchmark reflecting realistic deployment conditions, rather than as a fully controlled ablation. This is consistent with the primary contribution of this work, which is the introduction and characterization of MDWD as a benchmark dataset.

\section{Evaluation}
This section presents the comparative evaluation of all benchmark architectures on MDWD, reporting performance on both the validation and test partitions, with test-set results constituting the primary basis for comparison. Performance is assessed using $mAP_{50}$, $mAP_{50:95}$,  Precision, Recall, and F1-score.

\colorlet{bestvalcolor}{red!18}
\colorlet{besttestcolor}{cyan!18}
\begin{table*}[ht]
\centering
\caption{
Cross-architecture benchmark results on the \textbf{MDWD}, reporting validation (\textbf{V}) and test (\textbf{T}) performance for each metric.\\\textbf{RF-DETR-M} achieved the strongest overall test-set performance,
leading four of the five metrics.}
\label{tab:benchmark}
\footnotesize
\setlength{\tabcolsep}{1.5pt}
\renewcommand{\arraystretch}{1.08}
\begin{tabular*}{\textwidth}{@{\extracolsep{\fill}}
l
c
c
@{\hskip 2pt}c@{\hskip 0.5pt}c@{\hskip 0.5pt}
c@{\hskip 0.5pt}c@{\hskip 3pt}
c@{\hskip 0.5pt}c@{\hskip 3pt}
c@{\hskip 0.5pt}c@{\hskip 3pt}
c@{\hskip 0.5pt}c
}
\toprule
\textbf{Model} &
\textbf{Params} &
\textbf{Size} &
\multicolumn{2}{c}{\textbf{$\mathrm{mAP}_{50}$}} &
\multicolumn{2}{c}{\textbf{$\mathrm{mAP}_{50:95}$}} &
\multicolumn{2}{c}{\textbf{Precision}} &
\multicolumn{2}{c}{\textbf{Recall}} &
\multicolumn{2}{c}{\textbf{F1-score}} \\
&
\textbf{(M)} &
\textbf{(MB)} &
\textbf{V} & \textbf{T} &
\textbf{V} & \textbf{T} &
\textbf{V} & \textbf{T} &
\textbf{V} & \textbf{T} &
\textbf{V} & \textbf{T} \\
\midrule
YOLO26-N & 2.51 & 5.14 & 86.98 & 86.59 & 69.84 & 68.86 & 92.23 & 93.62 & 80.37 & 79.06 & 85.89 & 85.73 \\
YOLO26-S & 9.95 & 19.38 & 89.99 & 90.14 & 75.60 & 75.27 & 94.70 & 94.89 & 84.87 & 86.19 & 89.51 & 90.33 \\
YOLO26-M & 21.78 & 42.00 & 90.74 & 89.51 & 76.21 & 75.97 & 94.06 & 93.07 & 86.14 & 86.25 & 89.92 & 89.53 \\
YOLO26-L & 26.18 & 50.54 & 91.52 & 89.76 & 79.51 & 78.20 & 95.78 & \cellcolor{besttestcolor}\textbf{97.18} & 87.32 & 86.98 & 91.36 & 91.80 \\
\addlinespace[1pt]
\midrule
YOLO12-N & 2.57 & 5.26 & 88.35 & 88.01 & 71.22 & 70.29 & 91.55 & 94.02 & 82.16 & 80.39 & 86.60 & 86.67 \\
YOLO12-S & 9.26 & 18.06 & 89.41 & 90.00 & 74.14 & 74.41 & 92.69 & 95.72 & 84.58 & 85.13 & 88.45 & 90.11 \\
YOLO12-M & 20.14 & 38.88 & 91.47 & 90.30 & 76.56 & 75.84 & 94.16 & 95.72 & 87.59 & 85.73 & 90.76 & 90.45 \\
\addlinespace[1pt]
\midrule
YOLO11-N & 2.59 & 5.22 & 86.36 & 85.71 & 68.88 & 67.13 & 90.36 & 92.56 & 80.37 & 78.08 & 85.07 & 84.71 \\
YOLO11-S & 9.43 & 18.29 & 89.74 & 90.06 & 73.46 & 73.34 & 93.45 & 94.24 & 83.50 & 85.14 & 88.20 & 89.46 \\
YOLO11-M & 20.06 & 38.64 & 88.67 & 89.72 & 73.54 & 74.77 & 92.03 & 94.19 & 82.00 & 85.68 & 86.73 & 89.73 \\
\addlinespace[1pt]
\midrule
RF-DETR-N & 30.16 & 115.24 & 91.92 & 90.21 & 72.34 & 70.88 & 95.15 & 95.17 & 87.87 & 85.53 & 91.37 & 90.10 \\
RF-DETR-S & 31.81 & 121.52 & 94.27 & 92.61 & 77.38 & 76.17 & \cellcolor{bestvalcolor}\textbf{95.87} & 96.39 & 91.04 & 89.24 & 93.40 & 92.68 \\
\textbf{RF-DETR-M}
& 33.38
& 127.54
& \cellcolor{bestvalcolor}\textbf{95.81}
& \cellcolor{besttestcolor}\textbf{94.49}
& \cellcolor{bestvalcolor}\textbf{80.04}
& \cellcolor{besttestcolor}\textbf{78.21}
& 95.53
& 96.63
& \cellcolor{bestvalcolor}\textbf{91.70}
& \cellcolor{besttestcolor}\textbf{90.69}
& \cellcolor{bestvalcolor}\textbf{93.57}
& \cellcolor{besttestcolor}\textbf{93.56} \\
\bottomrule
\end{tabular*}
\vspace{2pt}
\par\footnotesize
\textit{Note:} The best \textbf{validation} result for each metric is highlighted in light red, while the best \textbf{test} result is highlighted in light blue.
\end{table*}

\subsection{Cross-Architecture Benchmark Results}
Table~\ref{tab:benchmark} reports validation and test performance across all evaluated architectures. Precision, recall, and F1-score are computed using a fixed-threshold matching procedure (confidence and IoU thresholds of 0.50), while $mAP_{50}$ and $mAP_{50:95}$ follow each framework's native COCO-style evaluation. RF-DETR-M achieves the strongest overall test-set performance, obtaining the highest values across all reported metrics, with RF-DETR-S providing closely comparable results at a slightly reduced model footprint. Within the YOLO family, YOLO26-L achieves the strongest overall performance, while smaller variants such as YOLO12-S and YOLO26-S remain competitive despite substantially lower parameter counts. All evaluated models achieve test-set F1-scores above 83\%, demonstrating that MDWD supports effective detector training across a broad range of architectural paradigms and model capacities. Furthermore, the relatively small validation-to-test performance differences observed across models indicate stable behavior on held-out imagery.

\subsection{Class-Level Analysis}
    Class-level results for RF-DETR-M reveal consistently strong performance across most categories on the test set. Recyclable Material achieves the highest performance (mAP\textsubscript{50} 98.3), followed by Organic Waste and Mixed Waste. Orange CMD exhibits high precision (96.9) but comparatively lower recall (79.5), indicating conservative predictions and a tendency to miss true instances. Other Waste is the most challenging category, with the lowest mAP\textsubscript{50} (93.2) and recall (83.5), and the highest false negative count (46), consistent with its heterogeneous, infrequently occurring waste types and greater intra-class variability.

\subsection{Discussion}
The strong performance achieved across convolutional, attention-centric, and transformer-based architectures indicates that MDWD is both learnable and sufficiently well annotated to support training across diverse detection paradigms. At the same time, the reduced performance observed for Orange CMD and particularly Other Waste demonstrates that the dataset remains challenging and provides meaningful headroom for future research on long-tailed and heterogeneous waste categories. The consistency between validation and test results further supports the reliability of the benchmark as a representative measure of generalization to unseen imagery. These results establish a reproducible baseline for future research on MDWD rather than asserting superiority of any individual architecture.

\section{Conclusion}
This paper introduced MDWD, a street-level benchmark for MSW detection comprising 3,697 images and 11,461 manually annotated instances across five operationally defined waste categories. By providing instance-level annotations within a structured municipal collection context, MDWD addresses a gap in existing waste-related datasets and establishes a reproducible benchmark for street-level waste monitoring. Experimental results demonstrate that the dataset supports effective training across both CNN-based and transformer-based detectors, with RF-DETR-M achieving the strongest overall performance.

While MDWD is centered on Malta's municipal collection context and currently focuses on bounding-box object detection, the benchmark provides a foundation for future research on more advanced perception paradigms. Future work may investigate promptable segmentation models such as SAM~3~\cite{hu2025sam3}, self-supervised visual representations such as V-JEPA~2~\cite{assran2025vjepa2}, and saliency-based prioritization mechanisms~\cite{dylansaliency} to support richer scene understanding, improved generalization, and operational decision-making in municipal monitoring workflows.
\noindent\rule{\linewidth}{0.3pt}

\bibliographystyle{IEEEtran}
\bibliography{refs}
\noindent\rule{\linewidth}{0.3pt}

\end{document}